\documentclass{article}
\usepackage{ijcai24}

\usepackage{times}
\usepackage{soul}
\usepackage{url}
\usepackage[hidelinks]{hyperref}
\usepackage[utf8]{inputenc}
\usepackage[small]{caption}
\usepackage{graphicx}
\usepackage{amsmath}
\usepackage{amsthm}
\usepackage{booktabs}
\usepackage[switch]{lineno}
\usepackage{caption}
\usepackage{subfigure}
\usepackage[table]{xcolor}
\usepackage{multirow}
\usepackage{booktabs}
\usepackage[ruled,linesnumbered]{algorithm2e}
\usepackage{algorithmic}
\title{Delocate: Detection and Localization for Deepfake Videos with Randomly-Located Tampered Traces}

\author{
Juan Hu$^1$
\and
Xin Liao$^1$\thanks{Corresponding author. }\and
Difei Gao$^{2}$\and
Satoshi Tsutsui$^{3}$\and
Qian Wang$^{4}$\and
Zheng Qin$^{1}$\and
Mike Zheng Shou$^{2}$\\
\affiliations
$^1$College of Computer Science and Electronic Engineering, Hunan University, China\\
$^2$Show Lab, National University of Singapore, Singapore\\
$^3$Rapid-Rich Object Search (ROSE) Lab, Nanyang Technological University, Singapore
\\
$^4$School of Cyber Science and Engineering, Wuhan University, China\\
\emails
\{hujuan, xinliao, zqin\}@hnu.edu.cn,
difei.gao@vipl.ict.ac.cn,
satoshi.tsutsui@ntu.edu.sg,
qianwang@whu.edu.cn,
mike.zheng.shou@gmail.com
}
\begin{document}

\maketitle

\begin{abstract}
Deepfake videos are becoming increasingly realistic, showing few tampering traces on facial areas that vary between frames. Consequently,  existing Deepfake detection methods struggle to detect unknown domain Deepfake videos while accurately locating the tampered region. To address this limitation, we propose \textsf{Delocate}, a novel Deepfake detection model that can both recognize and localize unknown domain Deepfake videos. Our method consists of two stages named recovering and localization. In the recovering stage, the model randomly masks regions of interest (ROIs) and reconstructs real faces without tampering traces, leading to a relatively good recovery effect for real faces and a poor recovery effect for fake faces. In the localization stage, the output of the recovery phase and the forgery ground truth mask serve as supervision to guide the forgery localization process. This process strategically emphasizes the recovery phase of fake faces with poor recovery, facilitating the localization of tampered regions. Our extensive experiments on four widely used benchmark datasets demonstrate that \textsf{Delocate} not only excels in localizing tampered areas but also enhances cross-domain detection performance.

\end{abstract}

\section{Introduction}\label{sec:intro}
Deepfakes, AI-generated videos of people, pose serious threats to society~\cite{chesney2019deep,aibase}, emphasizing the need for \textit{reliable} detection methods. By \textit{reliable}, we believe the following three characteristics are necessary: First, the method should be robust to unseen forgery patterns (Fig. \ref{fig1a}, \cite{ffdata,wild,dfdc,celeb}) with randomly located forgery traces (Fig. \ref{fig1b}), calling for cross-domain or cross-dataset evaluation. Second,  a truly \textit{reliable} method should convincingly explain the underlying reasons behind the model's decision by pointing to the manipulated part of a face. Unfortunately, we are not aware of works that satisfy all these criteria, so this paper develops a method that meets them.

\begin{figure}[!t]
 \centering
\includegraphics[width=\linewidth]{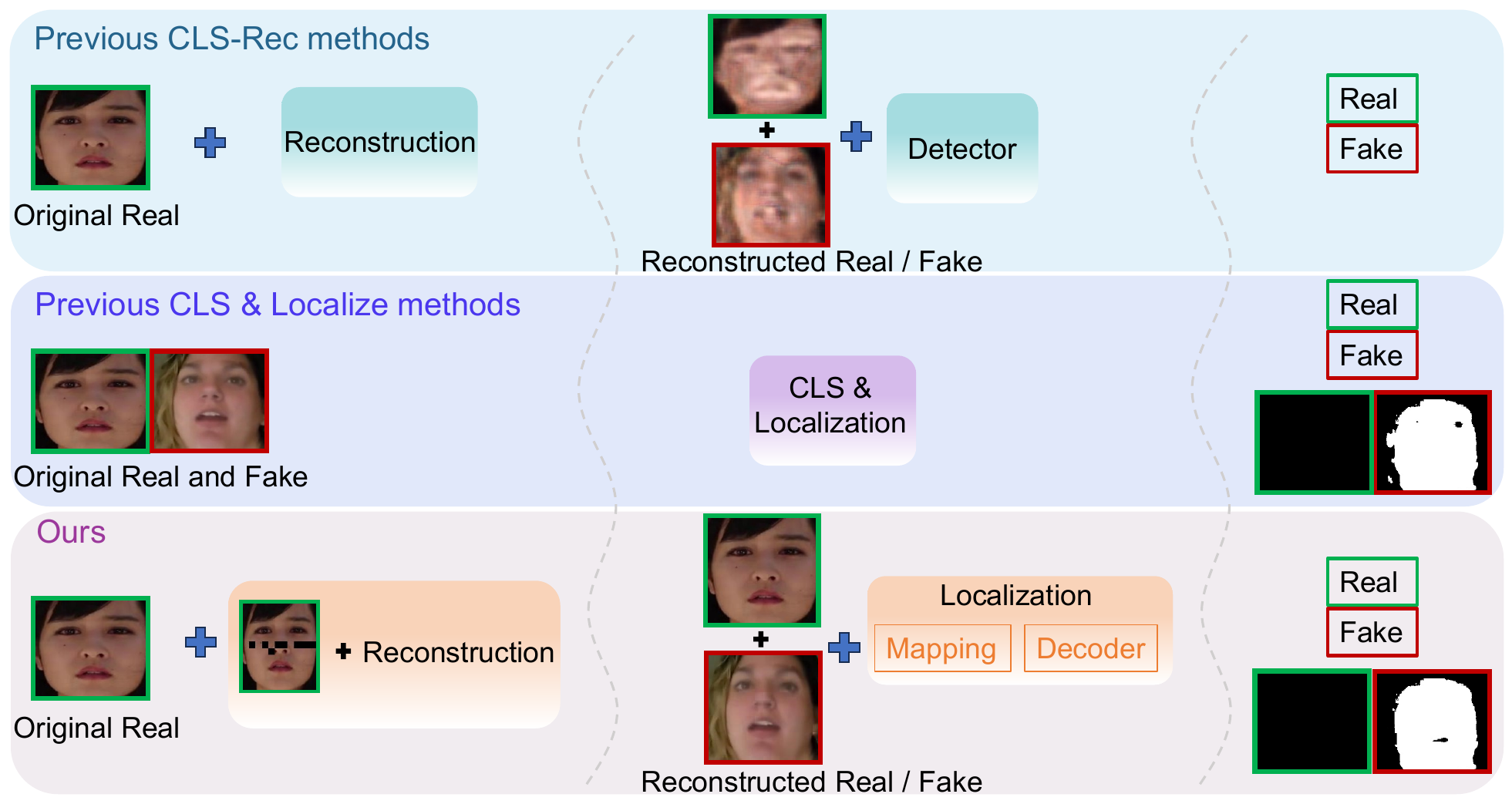}
   % \vspace{-0.3cm}
    \caption{Diferences between ours and previous methods. Previous CLS-Rec methods  mainly emphasize classification while overlooking localization aspects. Previous CLS \&  Localize methods leverage real and fake labels for feature extraction, without initially modeling real samples to extract robust features.  Our method integrates both classification and localization, with a dedicated focus on real samples, enabling us to extract features for enhanced performance.}
 \label{fig1}
% \vspace{-0.6cm}
\end{figure}

Recently, reconstruction-prediction-based methods have achieved relatively high detection performance. These methods typically involve the encoder and decoder that encode the input data into a low-dimensional representation and subsequently decode the original inputs from that representation. For example, \cite{khalid2020oc} uses reconstruction scores to classify real and fake videos. Moreover, reconstructing and predicting future frame representations \cite{finfer}, forgery configurations \cite{self}, pseudo training samples \cite{chen2022ost},  artifact representations \cite{dong2022explaining},  the whole faces \cite{recon,shi2023discrepancy}, the masked relation \cite{yang2023masked}, and mask regions \cite{chen2022learning} can boost the detection performance.
As illustrated in the first row of Fig. \ref{fig1}, since these methods typically train the reconstruction model solely with real images without specifically targeting any fake patterns, they have relatively good cross-domain performance. However, these methods overlook the importance of locating the forgery areas. Meanwhile, Deepfake localization methods\cite{he2021forgerynet,guo2023hierarchical,kong2022detect,lai2023detect,huang2022fakelocator,zhao2023hybrid,țanțaru2024weakly} can locate forgery areas, but they learn the representation directly using real and fake videos, leading to a performance drop in detecting unseen types of fakes.

\begin{figure*}[t]
  \centering
   \includegraphics[width=\linewidth]{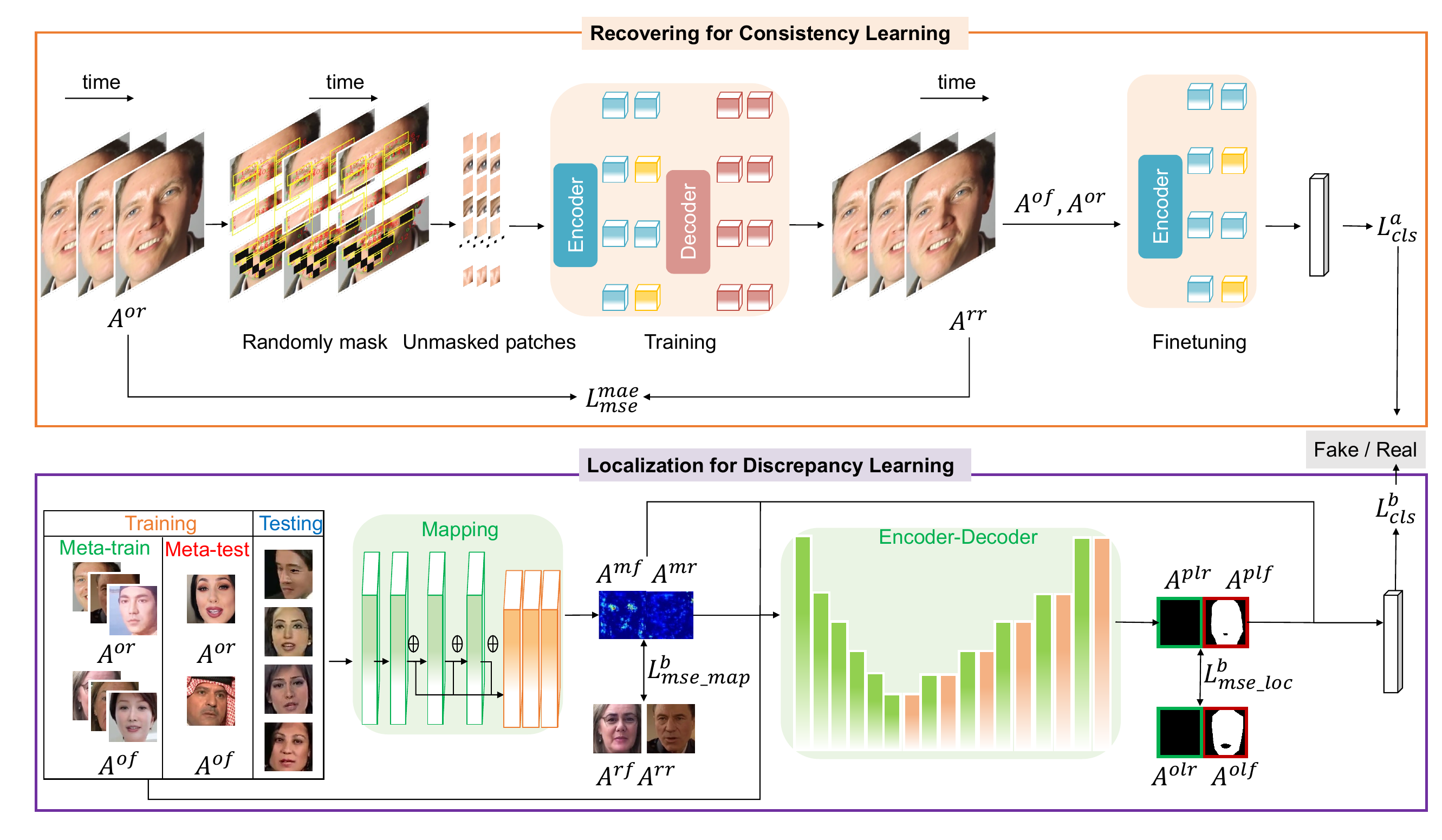}
% \vspace{-0.3cm}
   \caption{Pipeline of the proposed \textsf{Delocate}. In the Recovering stage, \textsf{Delocate} learns unspecific features by developing the designed masking strategy and recovery process. In the Localization stage, \textsf{Delocate} leverages devised mapping module and encoder-decoder module to maximize the discrepancy between real videos and Deepfake videos and locate the forgery areas.}
   %maximize the discrepancy between real videos and Deepfake videos, and then finally outputs the detection results.}
   \label{fig2}
% \vspace{-0.5cm}
\end{figure*}

In this paper, we design a method that (1) can robust to unseen forgery patterns with randomly located forgery traces and (2) can locate the manipulated parts of faces. We name our method as \textsf{Delocate}, which, in essence, works as follows. The first stage, \textit{Recovering for Consistency Learning} (Fig~\ref{fig2}-top), pretrains a masked autoencoder using real faces only. Training on real faces ensures that the method does not overfit to any Deepfake patterns, enabling better generalization to unseen generation techniques, addressing point (1). Moreover, to detect tampered traces that appear randomly on a face, we design a unique masking strategy guided by facial parts. Subsequently, the masked autoencoder predicts the masked regions of interest (ROIs) based on the unmasked facial parts and interframes. This strengthens the understanding of relationships between facial parts and their temporal consistency, addressing point (1). Furthermore, the second stage, \textit{Localization for Discrepancy Learning} (Fig~\ref{fig2}-bottom), combines meta-learning with localization supervision to explicitly enhance cross-dataset generalization performance while simultaneously localizing the tampered regions of fake faces, addressing points (1) and (2).

\noindent\textbf{Contributions}. 
\noindent{(1)} We propose \textsf{Delocate} to learn representations guided by facial parts, enabling the detection of Deepfake videos in unknown domains.

\noindent{(2)} Unlike most detection methods that simply predict real or fake, \textsf{Delocate} can precisely localize tampered regions on faces. Learning to localize actually enhances the model's ability to detect fake videos. 

% \noindent{(2)} To further explore general representation,  we utilize the reconstructed results and localization label for localization. By localizing the tampered regions, \textsf{Delocate} offers a solution to the pervasive issue of manipulation traces left in random areas by a range of Deepfake technologies. 

\noindent{(3)}  Extensive experiments on benchmark datasets, including  FaceForensics++ (FF++) \cite{ffdata}, Celeb-DF (CDF)  \cite{celeb}, DeeperForensics-1.0 (DFo) \cite{jiang2020deeperforensics}, DFDC \cite{dfdc} show that \textsf{Delocate} achieves effective performance under various metrics.

% \vspace{-0.2cm}
\section{Related Work}
%Before the emergence of the reconstruction-prediction-based methods explained in Sec.~\ref{sec:intro}, 
\textbf{Deepfake detection.}  Detection methods that focus on classifying real and fake videos can be broadly divided into two types:  classification based on generalized methods ({CLS-Gen}) and classification based on reconstruction-prediction methods ({CLS-Rec}). 
The generalized methods contain methods based on implicit clues, explicit clues, and both implicit and explicit clues. Methods that explore implicit clues~\cite{ffdata,smil,ltw,CNN-GRU,multiatt,aaaidual,aaailocal,pcl,dong2023implicit} use supervised learning to distinguish genuine and fake videos without explicitly incorporating clues to detect Deepfake videos, making it challenging to understand the underlying detection clues. Methods that employ explicit clues~\cite{eye,head,emotions,luo2021generalizing,on,lips,finfer,Patch-based,ftcn,delv,finfer,shiohara2022detecting,wang2023noise} %{eye,head,thinking,frequency,cnn2,emotions,luo2021generalizing,on,hcil,xray,lips,spsl,finfer,stil,UIA-ViT,Patch-based,pcl,ftcn,delv,haliassos2022leveraging,dong2022protecting,shiohara2022detecting,wang2023noise}
have achieved more promising performance. Furthermore, Huang et al. \cite{huang2023implicit} explore explicit and implicit embeddings for Deepfake detection. However, given the rapid advancement of Deepfake technology, various falsification traces can be left behind, rendering detection methods that rely on explicit features vulnerable to attack. 

The reconstruction-prediction-based methods \cite{khalid2020oc,finfer,self,chen2022ost,dong2022explaining,recon,yang2023masked,chen2022learning,shi2023discrepancy} are explained in Sec.~\ref{sec:intro}. Though these methods achieve promising detection performance, they do not focus on forgery localization. 

\noindent\textbf{Deepfake localization.} Kindly note that while there is a vast array of research papers on image localization, our discussion here is specifically focused on papers related to Deepfake classification and localization ({CLS \& Localize}). There are few works that focus on Deepfake localization. Kong et al. \cite{kong2022detect} use the noise map and semantic map to predict the forgery regions. Lai et al. \cite{lai2023detect} use the mask decoder to locate forgery areas and classify videos. Zhao et al. \cite{zhao2023hybrid} proposed RGB-Noise correlation to obtain the predicted manipulation regions. A recent paper \cite{shuai2023locate} proposes a two-stream network for improving detection performance. These methods push the Deepfake forensic one step further to forgery localization, but they struggle to classify cross-domain Deepfake videos.
% \vspace{-0.2cm}
\section{Method}
\label{sec:Method}
This section presents the details of \textsf{Delocate} for Deepfake video detection.  Specifically, the proposed method is composed of two stages: (1) Recovering for  Consistency Learning, and (2) Localization for  Discrepancy Learning stage, as shown in Fig.~\ref{fig2}. We demonstrate the logic design in Algorithm \ref{algorithm1}.

\begin{algorithm}[!t]
\caption{The algorithm process of \textsf{Delocate}. %The recovery process $\theta_\mathit{Rec}$, finetuning process $\theta_\mathit{Fin}$, and localization process $\theta_\mathit{Loc}$ are optimized by \texttt{AdamW}.
}
\label{algorithm1}

{\bf Input: } {\\The original real faces $A^{or}$, the original fake faces $A^{of}$. The original  localization $A^{ol}$ =\{$A^{olr}$,$A^{olf}$\}. The batch size $bs=8$. The number of iterations $num\_iter$. The learning rate $\alpha$.}

{\bf Output: } {\\Trained model in recovery  $\theta_{Rec}^{Ted}$, finetuning  $\theta_{Fin}^{Ted}$, and localization  $\theta_\mathit{Loc}^{Ted}$ process.}
\begin{algorithmic}[1] %[1] enables line numbers
\WHILE{(\textit{Recovery process}) \space $\theta_{Rec}$ have not converged }
{
\FOR {$i = 1 \to num\_iter^{Rec} $}
{\STATE
$A^{rr}=\theta_\mathit{Rec}(A^{or})$\\
\STATE $\theta^{grad}_{Rec} \gets \nabla_{\theta_\mathit{Rec}}(\frac{1}{b}\sum_{i=1}^{i=bs} {L^{mae}_{mse}}(A^{or}, A^{rr}) ) $\\
\STATE $ \theta_{Rec}^{Ted} \gets \theta_{Rec} - \alpha_\mathit{Rec} \cdot $ $AdamW$$(\theta_\mathit{Rec}, \theta^{grad}_{Rec}) $

}
\ENDFOR}\ENDWHILE  
% \space(\textit{Recovery process})
\WHILE{(\textit{Finetunig process}) \space $\theta_{Fin}$ have not converged}
{
\FOR {$i = 1 \to num\_iter^{Fin} $}
{\STATE
$p_{i}^{A^{o}\_Fin}=\theta_{Fin}(A^{or},A^{of})$\\
\STATE $\theta^{grad}_{Fin} \gets \nabla_{\theta_\mathit{Fin}}(\frac{1}{bs}\sum_{i=1}^{i=bs} {L^{a}_{cls}}(p_{i}^{A^{o}\_Fin}, y_{i}^{A^{o}\_Fin}) ) $\\
\STATE $ \theta_{Fin}^{Ted} \gets \theta_{Fin} - \alpha_\mathit{Fin} \cdot $ $AdamW$$(\theta_\mathit{Fin}, \theta^{grad}_{Fin}) $
}
\ENDFOR}\ENDWHILE 
% \space(\textit{Finetuning process})
\WHILE{(\textit{Localization process}) \space $\theta_{Loc}$ have not converged}
{
\FOR {$i = 1 \to num\_iter^{Loc} $}
{
\STATE
$A^{r}=\{A^{rr},A^{rf}\}=\theta_\mathit{Rec}^{Ted}(A^{or},A^{of})$\\
\STATE
$A^{m}=\{A^{mr},A^{mf}\}=\theta_{Loc}^{map}(A^{or},A^{of})$\\

\STATE
$p_{i}^{A^{o}\_Loc}, A^{pl} =\theta_{Loc}^{Cls}(A^{or},A^{of},A^{ol})$\\
\STATE $\theta^{map^{grad}}_{Loc} \gets \nabla_{\theta_\mathit{Loc}}(\frac{1}{bs}\sum_{i=1}^{i=bs} {L^{b}_{mse\_map}}(A^{m}, A^{r}) ) $\\
\STATE $\theta_{Loc}^{Cls^{grad}} \gets \nabla_{\theta_\mathit{Loc}}(\frac{1}{bs}\sum_{i=1}^{i=bs} {L^{b}_{cls}}(p_{i}^{A^{o}\_Loc}, y_{i}^{A^{o}\_Loc}) ) $\\
\STATE $\theta_{Loc}^{grad} \gets \nabla_{\theta_\mathit{Loc}}(\frac{1}{bs}\sum_{i=1}^{i=bs} {L^{b}_{mse\_loc}}(A^{ol},A^{pl}) ) $\\
\STATE $\theta_{Loc}^{Meta^{grad}} \gets \nabla_{\theta_\mathit{Loc}}(\frac{1}{bs}\sum_{i=1}^{i=bs} {L^{b}})$\\
%(\theta^{map^{grad}}_{Loc},\theta_{Loc}^{Cls^{grad}},\theta_{Loc}^{grad})} ) 
\STATE  {$ \theta_{Loc}^{Ted} \gets \theta_{Loc} - \alpha_\mathit{Loc} \cdot $ \\ $SGD$   $(\theta_\mathit{Loc}, \theta^{map^{grad}}_{Loc},\theta_{Loc}^{Cls^{grad}},\theta_{Loc}^{grad},\theta_{Loc}^{Meta^{grad}}) $}
}
\ENDFOR}\ENDWHILE 
% \space(\textit{Localization stage})
\end{algorithmic}
\end{algorithm}

\noindent\textbf{Notations. }
% Let $A^{or}=\{A^{or}_1,A^{or}_2,\cdots,A^{or}_n\}$ be $n$ original real faces. Let $A^{of}=\{A^{of}_1,A^{of}_2,\cdots,A^{of}_n\}$ be $n$ original fake faces. Let recovered real faces  be  $A^{rr}=\{A^{rr}_1,A^{rr}_2,\cdots,A^{rr}_n\}$. Let recovered fake faces  be  $A^{rf}=\{A^{rf}_1,A^{rf}_2,\cdots,A^{rf}_n\}$. Let masked real faces  be  $A^{mr}=\{A^{mr}_1,A^{mr}_2,\cdots,A^{mr}_n\}$. Let masked fake faces  be  $A^{mf}=\{A^{mf}_1,A^{mf}_2,\cdots,A^{mf}_n\}$.
Let $A^{or}$, $A^{of}$, $A^{rr}$, $A^{rf}$, $A^{mr}$, $A^{mf}$, $A^{olr}$, 
 $A^{olf}$, $A^{plr}$, $A^{plf}$ be original real faces, original fake faces, recovered real faces, recovered fake faces,  masked real faces, masked fake faces, original real face localization, original fake face localization, predicted real face localization, and predicted fake face localization.
\subsection{Recovering for Consistency Learning}\label{sec:mafpc}
In this stage, we perform self-supervised learning of real faces to learn generic facial part consistency features. As a result,  the unspecific inconsistencies of fake faces with randomly-located tampered traces are exposed. Furthermore, we finetune the model with real faces and fake faces.

\noindent \textbf{Masking strategy tailored to learn the consistent face representation. }We design a facial part masking strategy to ensure that the model can learn the consistencies of all facial parts. The designed facial part masking strategy is different from the frame masking strategy of VideoMAE \cite{videomae}.%We show the pseudo-code in Algorithm \ref{algorithm1}, which takes the inputs of videos' faces and outputs the masked faces. Following the temporal masking of VideoMAE \cite{videomae} and modifying the frame level masking strategy of VideoMAE \cite{videomae}, we enforce different frames share the same masking regions to ensure the mask expands over the entire temporal axis.

%Furthermore, since we split the faces into three sets and randomly select one set of patches to mask, the model can recover the masked area according to the other two sets, which prompts the model to focus on the consistencies among all three sets. 
% The original MAE~\cite{imgmae}  masks random patches of the input image, but the same strategy is not suitable for our \textsf{Delocate} for the following reasons. 

\begin{figure}[!t]
\centering
\subfigure[Different forgery patterns]
{
\includegraphics[width=4cm]{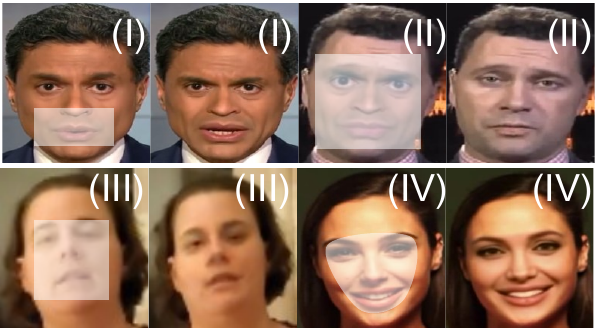}
\label{fig1a}
% \caption{fig4a}
}
\subfigure[Random forgery traces]
{
\includegraphics[width=4cm]{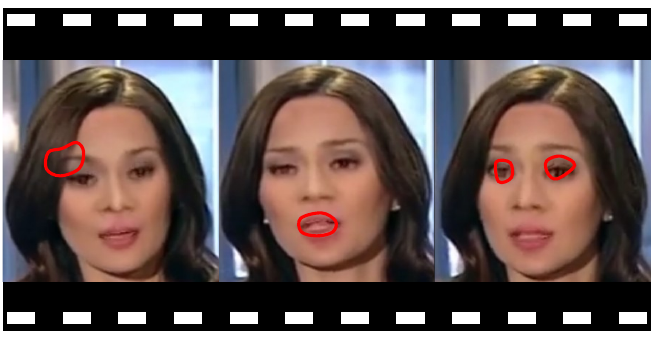}
\label{fig1b}
% \caption{fig4b}
}
% \vspace{-0.4cm}
\caption{The significance of the randomly-located traces. Different forgery patterns employ different shapes to alter the face area, rendering random tampered traces across different frames, which cannot be predicted based on the current frame, resulting in strong unpredictability. (I) Face2Face in FF++. (II) FSGAN in DFDC (III) DeepFakes in FF++. (IV) Deepfake in Celeb-DF.}
\label{figab}
% \vspace{-.2in}
\end{figure}

\indent  First,  as shown in Fig. \ref{fig1b},  the tampered traces may only be sporadically present in one part and not related to other facial parts. Hence, we devise the masking strategy by considering  Deepfake's domain knowledge. Specifically, we split the faces into different facial parts, i.e., eyes, cheek \& nose, and lips, enabling the model to focus on both local and global consistencies among all facial parts. We choose region-specific masking strategy instead of a haphazard approach because random masks can fail to maintain the crucial global consistency among various facial regions. Neglecting such global facial part consistency could impede the model's ability to learn accurate facial part consistency features, making it challenging to distinguish real from fake videos based on reconstructed faces. 
% {\renewcommand\baselinestretch{1.0}\selectfont

% \begin{algorithm}[!t]

% \caption{Masking process of facial parts}
% \label{algorithm1}

% \KwIn{Original faces: $A^{of}$, $A^{or}$,  Mask ratio: $M_r$;}

% \KwOut{Masked  faces: $A^{mf}$, $A^{mr}$ ;}

% Resize the face image with a size of 224*224, and detect $68$ keypoint landmarks in faces;

% Define $ROIs=\{R_1,R_2...R_{11}\}$;

% Define $196$ blocks of face image;

% %Partition image patches based on keypoints into three sets; 
% $M =\{M_1, M_2, M_3\} $, $M_1:$ eyes patches $\in$ \{$(0,0)$  $\rightarrow$  $17^{th}_{landmark}$\}, and $M_2:$ nose \& cheek patches $\in$ \{end of eyes patches   $\rightarrow$  $16^{th}_{landmark}$\}, and $M_3:$ lips patches $\in$ \{end of nose \& cheek patches   $\rightarrow$ the last patches of $A^{of}$, $A^{or}$\};

% %, which indicates patches of eyes, nose \& cheek, and lips region respectively, and random select one set of patches $M_{i}$;
%  \For {$i = 1$ \textbf{to} $3$}
%     {Randomly select $M_i$;
    
%      \For {$block \in selected$ \indent $M_{i}$}
%      {\If {$ROI$ \indent$landmarks \in block$}
%     {Sign the blocks $S_{blk}$; }
%     { Number of mask patches  = $S_{blk}$ $\times$ $M_r$;
     
%     Random mask the patches; }}}

% \end{algorithm}

% \par}
\indent Second, the original masking strategy of VideoMAE \cite{videomae}, with a high masking ratio, would make it too challenging to restore the original appearance without any artifacts or distortions. If reconstruction artifacts occur, real faces will contain them, and fake faces will display both reconstructed artifacts and tampering artifacts. This makes it difficult to distinguish real videos from fake videos since both have artifacts. Therefore, we propose a masking strategy that focuses on ROIs and utilizes a relatively low masking ratio to enable the model to reconstruct the original faces more accurately. 

The ROIs extraction is partially inspired by Facial Action Coding System (FACS) \cite{facs}, which considers the action units of FACS as fundamental elements. Drawing from psychology studies \cite{facs,mico1,micro2,expression,xinli}, it is well-known that real faces exhibit inherent consistency in these elements. Consequently, when we mask ROIs, it becomes more challenging to reconstruct these regions for fake faces compared to real faces.

We reference the action units of eyebrows,  lower eyelid, nose root, cheeks, mouth corner, side of the chin, and chin to calculate bounding box coordinates according to facial keypoints.  We discuss more about the masking strategy in the ablation study. \\
 \noindent \textbf{Network architecture.}  Our masked autoencoder is based on an asymmetric encoder-decoder architecture \cite{imgmae}. To consider temporal correlation, the vanilla  Vision Transformers (ViT)  and joint space-time attention \cite{videomae} are adopted for recovering.
 \\
   \textbf{Recover masked faces.} The masked patches of faces are dropped in the processing of the encoder, leaving the unmasked areas. In this way, the decoder predicts the missing facial part based on the unmasked areas. The reconstruction quality of masked patches is calculated with the MSE loss function $L^{mae}_{mse}$.
   \setlength{\abovedisplayskip}{0.2pt}
   \setlength{\belowdisplayskip}{1pt}
   \begin{equation}
L^{mae}_{mse}=\frac{1}{n} \sum_{i=1}^{n} (A^{or}_i - A^{rr}_i)^2.
\label{msemae}
\end{equation}
If the model learns consistencies among facial parts, the loss between the reconstructed patches and the input patches should decrease. Our facial part masking strategy makes each part selected randomly, which enforces the model to learn the representation unspecific to any facial part. Furthermore, since this phase solely utilizes authentic videos and excludes any Deepfake content, it helps prevent the model from overfitting to particular Deepfake tampering patterns. In this way, the pretrained recovery model is obtained. Let  $A^{rr} = RE^r \times A^{or}, \mbox{subject  to } 0<{RE}^r<1,$ where $RE^r$ represents the recovery quality of $A^{or}$.  A higher score of $RE^r$ indicates better reconstruction quality.

\noindent \textbf{Finetuning the recovery model. }We discard the decoder and apply the encoder to uncorrupted $A^{or}$  and  $A^{of}$  for finetuning. The finetuning process uses a cross-entropy loss for detection. 
\begin{equation}
L_{cls}^{a}= \frac{-1}{N} \sum_{i=1}^{N} \left[ y^{A^{o}}_i \log p^{A^{o}}_i + (1 - y^{A^{o}}_i) \log(1 - p^{A^{o}}_i)\right],
\label{clsa}
\end{equation}
where $p^{A^{o}}_i$ is the predicted label of original faces, $y^{A^{o}}_i$ is the groud truth label of original faces.  Since the recovery model learns the facial part consistency of real videos, the well-trained encoder can extract the consistency features of real videos. For fake videos, as shown in Fig. \ref{figab}, they are generated by different forgery patterns and tampered with different areas, and the tampered traces can show up in random regions. Consequently, the tampered traces can not be predicted. If the masked areas contain tampered traces, the recovery process would be affected. If there are no tampered traces in the masked area, the  tampered traces in unmasked areas can not be recovered. That is, regardless of whether the tampering traces are covered, the video with randomly-located tampering traces will influence the recovery process, which makes the features extracted from the encoder different from those of the original videos.

\subsection{Localization for Discrepancy Learning}\label{sec:stage2}
In this stage, we leverage the well-trained recovery model from the first stage and map the recovery result to enlarge the discrepancy between real and fake videos.

\noindent\textbf{Input data. }We load the trained recovery model to obtain $A^{or}$, and input  $A^{of}$, $A^{or}$, $A^{rf}$, and $A^{rr}$ into the Localization stage. Let $A^{rf} = RE^f \times A^{of}$, where $RE^f$ represents the recovery quality of $A^{of}$.

\noindent \textbf{Data split strategy.} To avoid over-fitting to specific Deepfake patterns, we use meta-learning \cite{meta} and randomly divide the training data into Meta-train set and Meta-test set, where fake faces in Meta-train and Meta-test have different manipulated patterns.  \\
   \textbf{Network architecture. } We utilize the first convolutional layer of ResNet-18 \cite{resnet}. Instead of directly utilizing ResNet-18, we employ the first three residual blocks of ResNet-18 and concat the outputs of these residual blocks. Second, the concatenated outputs are fed into three convolutional layers for face mapping. The dimensions of the mapped faces are $56 \times 56 \times 3$.  In this way, $A^{mr}$ and $A^{mf}$ can be represented as, $A^{mr} = MA^r \times A^{rr}, \mbox{subject  to } 0<MA^r<1$, $A^{mf} = MA^r \times A^{rf}, \mbox{subject  to } 0<MA^f<1$, where $MA^r$ and $MA^f$ represent the mapping quality of $A^{rr}$ and $A^{rf}$. 
Third, the extracted mapping features are leveraged for classification purposes. Lastly, these features, alongside the original faces and localization labels, are fed into an encoder-decoder framework. 

The encoder incorporates the SENet architecture \cite{hu2018squeeze}, while the decoder adopts the UNet framework \cite{ronneberger2015u}. To enhance the network's focus on pivotal regions, the SCSE Module \cite{wu2022robust} is integrated into the decoder. The classification outcomes derived from the mapping features are governed by the constraint $L_{mse\_map}^b$, establishing a link with the encoder-decoder's localization component. This localization module is similarly regulated by $L_{mse\_loc}^b$. Both the mapping and localization results collectively contribute to the overall classification constraint $L_{cls}^b$. Instead of focusing on one task of classification and localization, our classification and localization results are mutually constrained and mutually promoted to facilitate us to complete multi-tasks of classification and localization.
\noindent\textbf{Detection loss. } To amplify the differences between $A^{or}$ and $A^{of}$, we should satisfy:
\begin{equation}
A^{mr}-A^{mf} \gg A^{or}-A^{of}.
\label{mloss}
\end{equation}
Combine the analyses of the $A^{mr}$, $A^{mf}$, $A^{or}$, and $A^{of}$, Eq. (\ref{mloss}) can be represented as:
\begin{equation}
A^{rr}(MA^r - \frac{1}{RE^{r}}) \gg A^{rf}(MA^f - \frac{1}{RE^{f}}).
\label{cb1}
\end{equation}
% Since we have $0<RE^f<RE^r<1$ and $A^{rr}>A^{rf} $,  
Since the recovery model is trained on real data $A^{or}$ and the randomly-located traces of $A^{of}$ could influence the recovery process, we have $A^{rr}>A^{rf} $. Moreover, the recovery quality of $A^{of}$ can be smaller than that of $A^{or}$. That is, $0<RE^f<RE^r<1$. To satisfy Eq. (\ref{cb1}),  it is necessary to ensure that $MA^r \gg MA^f $. Therefore, we minimize the MSE loss $L_{mse\_map}^{b}$ between the mapped faces and the reconstructed faces. Consequently, $L_{mse\_map}^{b}$ allows the $A^{mr}$ to be constrained by the consistency of $A^{rr}$, while the $A^{rf}$ are constrained by the inconsistency. In this way, the model is able to recover $A^{rr}$ but fails to recover $A^{rf}$, ensuring that $MA^r \gg MA^f $. These discrepancies enable the model to detect the failed reconstructed faces and better locate the tampered areas without misjudging the real faces. 

For each pixel value in predicted localization masks $A^{plr}_{pix}$,  we normalize it and process it as follows. 
\begin{equation}
 A^{plr}_{pix}=\begin{cases} 
1, & \text{if } A^{plr}_{pix} \geq 0.5 \\
0, & \text{if } A^{plr}_{pix} < 0.5 
\end{cases}.
\label{cb2}
\end{equation}The primary objective of localization is to minimize the MSE loss $L_{mse\_loc}^{b}$ between the ground truth localization mask and predicted localization mask. 

Moreover, we also minimize the binary cross-entropy $L_{cls}^{b}$ between the video labels and the combined outputs of the mapping and localization features. 

For each epoch, a sample batch is formed with the same number of fake videos and real videos to construct the binary detection task. To simulate unknown domain detection during training, the Meta-train phase performs training by sampling many detection tasks, and is validated by sampling many similar detection tasks from the Meta-test. Thereafter, the parameters of Meta-train phase can be updated. The goal of Meta-test phase is to enforce a classifier that performs well on Meta-train and can quickly generalize to the unseen domains of Meta-test, so as to improve the cross-domain detection performance.

The final loss function of the Localization stage is:
\begin{equation}
\begin{split}
L^b &=  (  L^b_{cls} + L_{mse\_map}^{b} +L_{mse\_loc}^{b})_{Meta^{train}}+\\&( L^b_{cls}+ L_{mse\_map}^{b}+L_{mse\_loc}^{b})_{Meta^{train}} )_{Meta^{test}}  .
\end{split}
\end{equation}
which combines the Meta-test loss of $L_{cls}^{b}$, $L_{mse\_map}^{b}$, and $L_{mse\_loc}^{b}$ and Meta-train loss of $L_{cls}^{b}$, $L_{mse\_map}^{b}$, and $L_{mse\_loc}^{b}$ to achieve joint optimization.

\noindent\textbf{Detection results. }We average the output of Recovering stage and Localization stage to get the final detection score.

\section{Experiment}\label{sec:Experiment}
\subsection{Experimental Setup}
\indent \textbf{Datasets.} Four public Deepfake video datasets, i.e.,  FF++ \cite{ffdata}, CDF  \cite{celeb}, DFo \cite{jiang2020deeperforensics}, DFDC \cite{dfdc} are utilized to evaluate the proposed method and existing methods. FF++ is made up of $4$ types manipulated algorithms: DeepFakes (DF) \cite{deepfakegit},  Face2Face (F2F) \cite{thies2016face2face}, FaceSwap (FS) \cite{faceswapgit}, NeuralTextures (NT) \cite{ne}. Moreover, $4000$ videos are synthesized based on the $4$ algorithms. These videos are widely used in various Deepfake detection scenarios. Celeb-DF contains $5639$ videos that are generated by an improved DeepFakes algorithm \cite{celeb}. The tampered traces in some inchoate datasets are relieved in Celeb-DF. DeeperForensics-1.0  dataset is published for real-world Deepfake detection. DFDC  is a large-scale Deepfake detection dataset published by Facebook.

\noindent \textbf{Implementation details.} In the Recovering stage, the masking ratio, batch size, patch size, and input size are set as $0.75$, $8$, $16$, $224$, respectively. The AdamW \cite{adamw} optimizer with an initial learning rate $1.5 \times 10^{-4}$, momentum
of $0.9$ and a weight decay $0.05$ is utilized to train the recovery model. The finetuning of the Recovering stage utilizes the AdamW  optimizer with an initial learning rate $1 \times 10^{-3}$ to detect videos. The
SGD optimizer is used for optimizing the Localization stage with the initial learning rate $0.1$, momentum of $0.9$, and weight decay of $5 \times 10^{-4}$. We use \texttt{FFmpege} \cite{ffmpeg} to extract $30$ frames from each video. The \texttt{dlib} \cite{dlib} is utilized to extract  faces and detect $68$ facial landmarks. We follow Kong et al. \cite{kong2022detect} to extract the ground truth of forgery localization. %We randomly mask facial parts according to Algorithm \ref{algorithm1}. %Eyes and their surroundings begin at the coordinate $(0,0)$ and end at the coordinate of the $17^{th}$ landmark. Cheek \& nose areas start at the end of the aforementioned eyes areas and end at the $16^{th}$ landmark. The remaining area is the lips and its surroundings.

\noindent\textbf{Comparison methods. }%We consider the detection methods that have reproducible and published source codes so that we can conduct a comparative experiment in a fair way. Therefore, we
We compare \textsf{Delocate} with the {CLS-Gen} methods  that are representative of implicit methods, explicit methods, and explicit and implicit combined methods, i.e., MultiAtt \cite{multiatt}, LipForensics \cite{lips}, Huang et al. \cite{huang2023implicit}. We also compare \textsf{Delocate} with {CLS-Rec} methods, i.e.,  %FInfer \cite{finfer}, SSL-AE \cite{self}, 
OST \cite{chen2022ost},   RECCE \cite{recon},  MRL \cite{yang2023masked}, and DisGRL \cite{shi2023discrepancy}.
 Furthermore, we compare the {CLS \& Localize} methods, i.e., Kong et al. \cite{kong2022detect}, Zhao et al. \cite{zhao2023hybrid}, Chao et al. \cite{shuai2023locate} .%We conducted a comparative experiment in a fair way

\begin{table}[!t]
\begin{center}
% \vspace{-0.4cm}

\renewcommand{\tabcolsep}{0.7mm} % enlarge column spacing
\renewcommand{\arraystretch}{1.0}
\begin{tabular}{ccccccc}
\noalign{\hrule height 0.7pt}

     % \multirow{2}{*}{Method}&\multicolumn{4}{c}{Testing datasets}\\ \cline{2-5}
     % & {Celeb-DF}&{WildDF}& {DFDC}&{Avg}\\

\multirow{2}{*}{Method}&\multicolumn{2}{c} {CDF}&\multicolumn{2}{c} {DFo}& \multicolumn{2}{c} {DFDC}\\ 
    \cmidrule(r){2-3} \cmidrule(r){4-5}\cmidrule(r){6-7}&AUC $\uparrow$ &EER$\downarrow$ &AUC $\uparrow$ &EER$\downarrow$&AUC $\uparrow$ &EER$\downarrow$\\
     
              \noalign{\hrule height 0.7pt}

% \textmd{FWA \cite{fwa}}&$69.5$& $68.2$&$67.3$&$68.3$\\

% {\textmd{OC-FakeDect}}&$65.8$&$38.3$&$66.4$&$39.2$& $69.6$&$35.8$&$67.3$&$37.8$ \\
{\textmd{MultiAtt}}&$76.7$&$32.8$&$72.4$&$34.7$& $67.3$&$38.3$\\
{\textmd{LipForensics}}&$82.4$&$24.2$&$97.6$&$10.6$& $73.5$&$36.5$\\
{\textmd{Huang et al.}}&$83.8$&$24.9$&$90.8$&$15.3$& $81.2$&$26.8$\\
\hline
% \multirow{5}{*}{CLS-Reconstruct}&{\textmd{FInfer}}&$70.6$&$34.8$&$\--$&$86.8$&$22.0$&$\--$& $70.9$&$36.4$&$\--$\\
% \multirow{4}{*}{CLS-Reconstruct}&{\textmd{SSL-AE}}&$79.7$&$30.8$&$\--$&$86.1$&$22.8$&$\--$& $72.3$&$35.8$&$\--$ \\
{\textmd{OST }}&$74.8$&$31.2$&$95.1$&$9.7$& ${83.3}$&$25.0$ \\

% {\textmd{FST-Matching }}&$89.4$&$18.2$&$78.3$&$29.4$& ${80.1}$&$26.0$&$82.6$&$24.5$ \\
{\textmd{RECCE}}&$73.7$&$30.3$&$89.3$&$16.9$& $74.0$&$31.1$ \\

{\textmd{MRL}}&$86.7$&$18.3$&$91.1$&$15.6$& $74.5$&$30.1$ \\
{\textmd{DisGRL}}&$76.7$&$28.3$&$88.4$&$18.5$& $74.8$&$30.0$ \\
\hline
{\textmd{Kong et al.}}&$70.7$&$35.5$&$82.6$&$24.7$& $63.3$&$40.8$\\
{\textmd{Zhao et al.}}&$74.8$&$30.0$&$80.9$&$25.8$& $79.0$&$26.1$\\
{\textmd{Chao et al.}}&$86.2$&$18.1$&${99.0}$&$7.6$& $82.5$&$25.1$\\

% {\textmd{Chen et al.}}&$82.8$&$21.2$&$76.5$&$30.0$&$73.2$&$31.7$&$77.5$&$27.3$ \\
\rowcolor{gray!20}
{\textsf{Delocate}}
&${\textbf{91.3}}$&$\textbf{{14.1}}$&$\textbf{{99.1}}$&${\textbf{6.6}}$&$\textbf{{84.0}}$&$\textbf{24.7}$\\

\noalign{\hrule height 0.7pt}

\end{tabular}
\end{center}
\caption{Comparisons of detection performance  (AUC (\%) and EER (\%)) between \textsf{Delocate} and other methods on CDF, DFo, and DFDC datasets when trained on $4$ types of videos of FF++. }
% \vspace{-0.3cm}
\label{crossdatasets}
% \vspace{-0.8cm}
\end{table}

\subsection{Generalization to Unknown Domains}
 We enforce \textsf{Delocate} to learn unspecific features for Deepfake video detection with randomly-located tampered traces. The unknown domain detection is precisely the scenario where tampered traces are often randomly-located. To test the performance of  \textsf{Delocate}, we simulate unknown domain Deepfake detection in multiple scenarios.

\textbf{Comparisons of classification. }First, we conduct experiments by training the model on FF++ with all $4$ types of videos, but testing on other datasets, i.e., CDF, DFo, DFDC, and we use Area Under Curve (AUC) and Equal Error Rate (EER) to evaluate the classification performance. The enormous differences between the training domain and the testing domain make it challenging to improve unknown domain detection performance. Nonetheless, the results in Table \ref{crossdatasets} show that  \textsf{Delocate} manages to improve the classification performance and achieve comparable localization performance at the same time. For example, \textsf{Delocate} improves the AUC on CDF from
86.2\% (the localization method: Chao et al. \cite{shuai2023locate}) to 91.3\%.%reach slight advantages in terms of average AUC and EER.

\begin{table*}[!t]
\begin{center}

\renewcommand{\tabcolsep}{2mm} % enlarge column spacing
\renewcommand{\arraystretch}{1.0}
\begin{tabular}{ccccccccccccc}
\noalign{\hrule height 0.7pt}
\multirow{2}{*}{Method}&\multicolumn{3}{c} {DF}&\multicolumn{3}{c} {F2F}& \multicolumn{3}{c} {FS}&\multicolumn{3}{c} {NT}\\ 
    \cmidrule(r){2-4} \cmidrule(r){5-7}\cmidrule(r){8-10}\cmidrule(r){11-13}&CDF &DFo &DFDC &CDF &DFo &DFDC&CDF &DFo &DFDC&CDF &DFo &DFDC\\
     
              \noalign{\hrule height 0.7pt}

% \textmd{FWA \cite{fwa}}&$69.5$& $68.2$&$67.3$&$68.3$\\

% {\textmd{OC-FakeDect}}&$65.8$&$38.3$&$66.4$&$39.2$& $69.6$&$35.8$&$67.3$&$37.8$ \\
{\textmd{MultiAtt}}&$68.7$&$80.6$&$70.1$&$69.6$& $81.9$&$68.6$&$70.4$&$82.5$&$70.1$&$70.2$&$82.9$&$66.9$ \\
{\textmd{LipForensics}}&$69.3$&$90.1$&$70.8$&$69.1$& $72.4$&$71.4$&$72.3$&$71.9$&$71.8$&$70.9$&$73.2$&$69.8$ \\
{\textmd{Huang et al.}}&$72.9$&$90.9$&$72.8$&$74.2$& $91.2$&$75.8$&$72.7$&$89.9$&$71.9$&$74.8$&$91.3$ &$73.5$\\
\hline
% \multirow{5}{*}{CLS-Reconstruct}&{\textmd{FInfer}}&$68.6$&$89.9$&$67.3$&$68.1$& $88.3$&$66.4$&$70.6$&$84.6$&$67.7$&$70.7$&$86.4$&$69.7$\\
% \multirow{4}{*}{CLS-Reconstruct}&{\textmd{SSL-AE}}&$73.0$&$74.2$&$\textbf{77.2}$&$78.1$& $78.6$&$78.7$&$80.0$&$77.2$&$69.5$&$75.9$&${75.5}$&$74.1$ \\
{\textmd{OST}}&${76.6}$&${93.8}$&${75.7}$&$79.9$& $\textbf{94.7}$&${79.8}$&$79.2$&$90.9$&${80.2}$&$75.3$&$\textbf{92.9}$&$75.2$\\

% {\textmd{FST-Matching }}&$89.4$&$18.2$&$78.3$&$29.4$& ${80.1}$&$26.0$&$82.6$&$24.5$ \\
{\textmd{RECCE}}&$69.7$&$78.0$&$68.0$&$70.5$& $75.7$&$71.1$&$69.7$&$73.3$&$71.1$&$70.1$&$74.5$&$70.2$\\
{\textmd{MRL}}&$72.9$&$79.3$&$72.2$&$70.6$& $79.5$&$71.2$&$73.1$&$84.2$&$70.5$&$71.4$&$82.1$&$72.4$\\
{\textmd{DisGRL}}&$71.5$&$79.2$&$70.2$&$70.3$& $78.9$&$72.0$&$73.3$&$82.9$&$71.0$&$72.8$&$83.7$&$72.3$\\
% {\textmd{Chen et al.}}&$82.8$&$21.2$&$76.5$&$30.0$&$73.2$&$31.7$&$77.5$&$27.3$ \\
\hline
{\textmd{Kong et al.}}&$69.3$&$80.8$&$62.6$&$68.4$& $79.4$&$62.1$&$69.2$&$79.2$&$62.9$&$70.1$&$79.2$&$62.3$\\
{\textmd{Zhao et al.}}&$71.2$&$79.8$&$76.2$&$70.4$& $79.6$&$76.1$&$73.0$&$79.0$&$75.9$&$71.8$&$79.2$&$74.6$\\
{\textmd{Chao et al.}}&$72.4$&$89.1$&$75.0$&$79.7$& $90.6$&$76.2$&$80.4$&$90.5$&$80.1$&$75.4$&$91.6$&$72.3$\\
\rowcolor{gray!20}
{\textsf{Delocate}}
&$\textbf{78.2}$&$\textbf{94.5}$&$\textbf{76.3}$&$\textbf{80.9}$& ${93.3}$&$\textbf{79.9}$&$\textbf{81.6}$&$\textbf{91.5}$&$\textbf{80.8}$&$\textbf{76.8}$&${90.9}$&$\textbf{75.9}$\\

\noalign{\hrule height 0.7pt}
% \vspace{-1.2cm}
\end{tabular}
\end{center}
\caption{Comparisons of the detection performance  (AUC (\%)) between \textsf{Delocate} and other methods on CDF, DFo, and DFDC datasets when trained on one type of videos of FF++. }
% \vspace{-0.3cm}
\label{crossffone}
\end{table*}

Second, to avoid performing experiments on a particular training mode, we change the training mode and conduct other unknown domain detection experiments. Specifically, we implement experiments by selecting one type of FF++ for training, but testing on other datasets, i.e., CDF, DFo, DFDC. Since there is only one type of video for training in experiments, we randomly split the training data into Meta-train and Meta-test with $7:3$. Results in Table \ref{crossffone} illustrate that \textsf{Delocate} outperforms previous methods in many scenarios. Compared with classification methods,   OST \cite{chen2022ost} performs better than \textsf{Delocate} in $3$ scenarios. Despite these results, it is worth noting that \textsf{Delocate}  achieves better classification performance, especially with a 2.4\% improvement over OST \cite{chen2022ost} when training on FS and testing on CDF. We also observe that  \textsf{Delocate}  performs better AUC performance than that  of localization methods  \cite{kong2022detect,zhao2023hybrid,shuai2023locate}. For instance, when training on DF and testing on CDF, \textsf{Delocate} achieves a 5.8\% AUC improvement over Chao et al. \cite{shuai2023locate}.

\textbf{Comparisons of localization. }
We use Intersection over Union (IoU) and Pixel-wise Binary Classification Accuracy (PBCA) \cite{kong2022detect} to evaluate the localization performance. We train the model on FF++ and test it in other datasets. Table \ref{iou} shows that Chao et al. \cite{shuai2023locate} the best IoU results in testing DFDC. \textsf{Delocate} performs best results in other scenarios. 

\begin{figure}[t]
 \centering
\includegraphics[width=\linewidth]{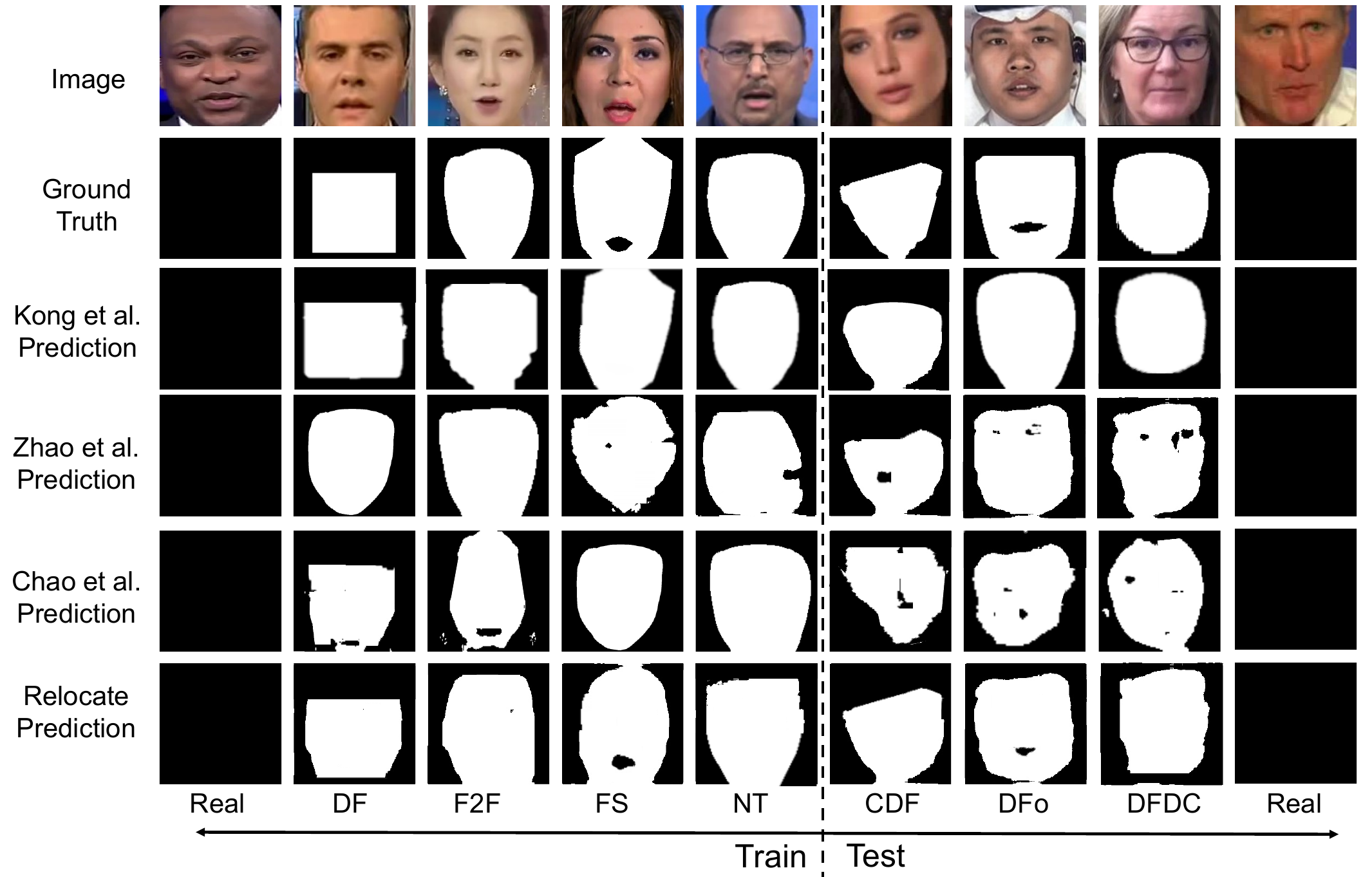}
   % \vspace{-0.7cm}
\caption{Comparisons of predicted forgery regions on CDF, DFo, and DFDC datasets when trained on $4$ types of videos of FF++. }
 \label{fig3mask}
% \vspace{-0.5cm}
\end{figure}

We also conduct forgery localization analyses for the CLS \& Localize methods and show the results in  Fig. \ref{fig3mask}. It shows that the localization area identified by Kong et al. \cite{kong2022detect}, Zhao et al. \cite{zhao2023hybrid} and Chao et al. \cite{shuai2023locate} exhibits sporadic mismatches across various regions when compared to the ground truth.  For the CDF, DFo, and DFDC datasets, \textsf{Delocate} aligns more closely with the ground truth region compared to the area localized by Zhao et al. \cite{zhao2023hybrid} and Chao et al. \cite{shuai2023locate}. It may be because \textsf{Delocate} focuses on unspecific features during the reconstruction stage, thereby revealing inconsistencies in the synthetic faces. In the localization stage, it maps the outcomes of the reconstruction, where the classification and localization results mutually influence and enhance each other. This process leads to the extraction of more generalized features, consequently improving the cross-domain performance.
\begin{table}[!t]
\setlength{\abovecaptionskip}{0.1cm}
\setlength{\belowcaptionskip}{0.1cm}
\tabcolsep=2pt
\begin{center}

\renewcommand{\tabcolsep}{1mm} % enlarge column spacing
\renewcommand{\arraystretch}{1.0}
\begin{tabular}{ccccccc}
\noalign{\hrule height 0.7pt}
\multirow{2}{*}{Method}&\multicolumn{2}{c} {CDF}&\multicolumn{2}{c} {DFo}& \multicolumn{2}{c} {DFDC}\\ 
    \cmidrule(r){2-3} \cmidrule(r){4-5}\cmidrule(r){6-7}&IoU $\uparrow$ &PBCA$\uparrow$ &IoU $\uparrow$ &PBCA$\uparrow$&IoU $\uparrow$ &PBCA$\uparrow$\\
           
\noalign{\hrule height 0.7pt}
\textmd{Kong et al.}&$0.709$&$0.721$ &$0.843$ &$0.826$&$0.616$&$0.624$ \\
 
\textmd{Zhao et al.}&$0.789$&$0.767$ &$0.904$ &$0.905$&$0.708$&$0.706$ \\

 \textmd{Chao et al.}&$0.798$&$0.784$ &$0.921$ &$0.919$&$\textbf{0.741}$&$0.726$ \\
  \rowcolor{gray!20}
\textmd{\textsf{Delocate}}&${\textbf{0.801}}$&${\textbf{0.802}}$ &${\textbf{0.937}}$ &${\textbf{0.926}}$ &${{0.738}}$ &${\textbf{0.727}}$  \\

\noalign{\hrule height 0.7pt}

\end{tabular}
\end{center}
\caption{Comparisons of localization performance (IoU and PBCA)  between \textsf{Delocate} and localization methods on CDF, DFo, and DFDC datasets when trained on $4$ types of videos of FF++. }
\label{iou}
% \vspace{-0.5cm}
\end{table}
\subsection{Intra-dataset Detection Performance} To provide a comprehensive assessment of the proposed \textsf{Delocate}, we compare \textsf{Delocate} with the state-of-the-art methods in the scenario of intra-dataset detection. Specifically, we conduct experiments on $4$ subsets of FF++ 
 (C23). The training data and testing data of intra-dataset experiments are from the same subset of FF++. Table \ref{intra} shows that most methods perform well in intra-dataset detection.  Chao et al. \cite{shuai2023locate} achieves the highest intra-dataset detection score while \textsf{Delocate} has a slight decrease of  $0.2\%$ in average accuracy compared. This drop may be due to the fact that the model improves the unknown domain performance while sacrificing a little bit of intra-domain performance to fit the unseen domain. %In terms of cross-manipulation detection performance, as shown in Table \ref{crossffone}, \textsf{Delocate} surpasses LipForensics\cite{lips} in multiple metrics.

\begin{table}[!t]
% \vspace{-0.4cm}
\setlength{\abovecaptionskip}{0.1cm}
\setlength{\belowcaptionskip}{0.1cm}
\tabcolsep=2pt
\begin{center}

\renewcommand{\tabcolsep}{4mm} % enlarge column spacing
\renewcommand{\arraystretch}{1.0}
\begin{tabular}{cccccc}
\noalign{\hrule height 0.7pt}
      {Method}& {DF}&{FS}& {F2F}&{NT}\\
           
\noalign{\hrule height 0.7pt}

% {\textmd{Xception \cite{ffdata}}}&$98.9$& $99.6$&$98.9$ &$95.0$&$98.1$ \\
 
% {\textmd{ADDNet \cite{wild}}}&$92.1$&$92.5$& $83.9$ & $78.2$&$86.7$\\
 
% {\textmd{S-MIL \cite{smil}}}&$98.6$&${99.3}$& $99.3$ & $95.7$&$98.2$\\
 
% {\textmd{STIL \cite{stil}}} & $99.6$&$\textbf{100}$& ${99.3}$& ${95.4}$& ${98.6}$\\
 
% {\textmd{HCIL \cite{hcil}}}&${\textbf{100}}$&${\textbf{100}}$& ${99.3}$& ${96.8}$&${99.0}$ \\

% {\textmd{STDT \cite{transformer22mm}}}&${99.8}$&${99.9}$& ${99.1}$& ${\textbf{98.0}}$&$\textbf{99.2}$ \\
{\textmd{MultiAtt}}&$99.6$&${\textbf{100}}$& ${99.3}$& ${98.3}$\\
{\textmd{LipForensics}}&${{99.8}}$&${\textbf{100}}$& $99.3$ & ${\textbf{99.7}}$\\
{\textmd{Huang et al. }}&$99.6$&${{99.8}}$& $99.5$ & ${98.4}$\\
\hline
% \multirow{5}{*}{CLS-Reconstruct}&{\textmd{FInfer }}&$98.4$&${96.0}$& ${93.5}$ & $94.9$\\
% \multirow{4}{*}{CLS-Reconstruct}&{\textmd{SSL-AE }}&${99.2}$&${99.4}$& $96.0$ & $99.0$\\
{\textmd{OST}}&${{99.0}}$&${98.8}$& $99.1$ & $95.9$\\
{\textmd{RECCE }}&$99.7$&${{99.9}}$& $99.2$ & ${98.4}$\\
{\textmd{MRL }}&${{99.2}}$&${{98.1}}$& $97.3$ &$98.6$\\
{\textmd{DisGR}}&${{99.0}}$&${{99.1}}$& $98.3$ &$99.6$\\
\hline
{\textmd{Kong et al.}}&$99.7$&${{99.6}}$& $99.4$ & ${98.9}$\\
{\textmd{Zhao et al.}}&$99.8$&${{99.4}}$& $99.0$ & ${97.9}$\\
{\textmd{Chao et al. }}&$\textbf{{100}}$&$\textbf{{100}}$& $\textbf{99.9}$ &$99.4$\\
% {\textmd{STIL \cite{stil}}} & $99.6$&${\textbf{100}}$& ${99.3}$& ${95.4}$& ${98.6}$\\
% {\textmd{ADDNet \cite{wild}}}&$92.1$&$92.5$& $83.9$ & $78.2$&$86.7$\\
% {\textmd{HCIL \cite{hcil}}}&${\textbf{100}}$&${\textbf{100}}$& ${99.3}$& ${96.8}$&${99.0}$ \\

 % {\textmd{PCL \cite{pcl}}}& ${\textbf{100}}$&${\textbf{100}}$&${\textbf{99.6}}$& $99.6$&${\textbf{99.8}}$\\

  \rowcolor{gray!20}
{\textsf{Delocate}}
&${99.8}$&$99.7$&${{99.7}}$&$99.3$\\

\noalign{\hrule height 0.7pt}
\end{tabular}
\end{center}
\caption{Comparisons of the Intra-dataset evaluation  (AUC (\%)) between \textsf{Delocate} and other methods.}
\label{intra}
% \vspace{-0.6cm}
\end{table}

% \begin{table}
% \setlength{\abovecaptionskip}{0.1cm}
% \setlength{\belowcaptionskip}{0.1cm}
% \tabcolsep=2pt
% \begin{center}
% \caption{ AUC (\%) scores of the FF++ testset  that are subject to different processing operations.}
% \label{operation}
% \renewcommand{\tabcolsep}{1.5mm} % enlarge column spacing
% \renewcommand{\arraystretch}{1.0}
% \begin{tabular}{ccccc}
% \noalign{\hrule height 0.7pt}
%      {Method}& {Resizing}& {Enhancement}&{Brightness}&{Contrast}\\
           
% \noalign{\hrule height 0.7pt}
%  {\textmd{R34+SBI}}& ${93.5}$& ${87.5}$&$83.2$&$84.5$ \\
%  {\textmd{EB4+SBI}}& ${97.1}$& ${90.8}$&$87.6$&$87.4$ \\

%   \rowcolor{gray!20}
% {\textsf{Delocate}}&${\textbf{97.8}}$&${\textbf{91.0}}$ &${\textbf{89.9}}$ &${\textbf{90.3}}$ \\
% \noalign{\hrule height 0.7pt}

% \end{tabular}
% \end{center}
% \vspace{-0.5cm}
% \end{table}

\begin{table}[!t]
\setlength{\abovecaptionskip}{0.1cm}
\setlength{\belowcaptionskip}{0.1cm}
\tabcolsep=2pt
\begin{center}

\renewcommand{\tabcolsep}{4mm} % enlarge column spacing
\renewcommand{\arraystretch}{1.0}
\begin{tabular}{cccc}
\noalign{\hrule height 0.7pt}
     {Mask ratio}& {CDF}& {DFo}&{DFDC}\\
           
\noalign{\hrule height 0.7pt}
\textmd{55\%}&$89.0$&${94.2}$ &$80.8$  \\
 
\textmd{65\%}&$90.6$&${92.6}$ &$81.8$  \\
  % \rowcolor{gray!20}
\textmd{75\%}&${\textbf{91.3}}$&${\textbf{99.1}}$ &${\textbf{84.0}}$  \\
 
\textmd{85\%}&$90.3$&${92.8}$ &$81.9$  \\
 
 \textmd{95\%}&$89.9$&${92.7}$ &$81.0$  \\
\noalign{\hrule height 0.7pt}

\end{tabular}
\end{center}
\caption{Ablation study - The detection performance (AUC (\%)) of different masking ratios on testing datasets after training on FF++.}
\label{ratio}
% \vspace{-0.7cm}
\end{table}

% \subsection{Robustness to  Post-Processing Operations}
% In the real-world situation, the frames and videos are often post-processed to adjust the media content for better display. The operations such as image resizing, image enhancement, video brightness, and video contrast are widely used. We post-precessing the frames or videos by using these operations and show the performance in Table \ref{operation}. After post-processing, specific forgery clues are relieved, thus SBI\cite{shiohara2022detecting} demonstrates sub-optimal results. While our method considers the unspecific facial part consistency, thus maintaining better detection performance even after post-processing.
% \vspace{-0.1cm}
\subsection{Ablation Study}
\label{ablation}

% To implement the ablation study, we conduct experiments by training on FF++ but testing on CDF, DFo, and DFDC datasets. 
% 

We conduct  ablation study experiments by training on FF++ but testing on CDF, DFo, and DFDC datasets. 

\noindent\textbf{Influence of the masking ratio.} We trained models on the FF++ dataset with different masking ratios. Note that instead of defining the masking ratio as the ratio of masked area to the entire face, we define the masking ratio as the ratio of masked area to the corresponding ROIs facial parts.  We choose not to use the original definition of mask ratio, which measures the ratio of the mask area to the entire face. Instead, we focus on specific regions of interest (ROIs) and divide the face into three parts. Then, we randomly mask only one part at a time.  Our attention is directed towards the specific masked ROIs during the masking procedure, rather than considering the entire face as a whole. 

\indent In Table \ref{ratio}, we observe that \textsf{Delocate} scales well with the masking ratio of 75\%.  The performance gets a slight drop in the masking ratio of 55\% and 65\% indicating that low masking ratios may hinder learning robust features. When the mask rate is 85\% and 95\%, the detection performance is also degraded. That may be because that high masking ratio can raise the difficulty of reconstructing faces. If both real faces and fake faces are not reconstructed well, the distinction between them can be reduced. Therefore, we set the masking ratio as 75\%  in the experiments.

\begin{table}
\setlength{\abovecaptionskip}{0.1cm}
\setlength{\belowcaptionskip}{0.1cm}
\tabcolsep=2pt
\begin{center}

\renewcommand{\tabcolsep}{3mm} % enlarge column spacing
\renewcommand{\arraystretch}{1.0}
\begin{tabular}{cccc}
\noalign{\hrule height 0.7pt}
     {Masking strategy}& {CDF}& {DFo}&{DFDC}\\
           
\noalign{\hrule height 0.7pt}

\textmd{MAE masking}&$86.4$&$95.8$&$79.1$  \\
 \textmd{VideoMAE masking}&$86.5$&$95.6$&$79.5$  \\
\textmd{Eye}&$91.1$&$98.7$& $80.1$\\

{\textmd{ cheek \& nose}}&$90.2$&$88.2$& $81.3$ \\

{\textmd{Lip}}&$90.8$&$88.2$& $81.7$ \\

 {\textmd{w/o ROIs}}&$90.9$&$88.9$& $83.5$ \\
 
 \rowcolor{gray!20}
 {\textmd{Proposed strategy}}&${\textbf{91.3}}$&${\textbf{99.1}}$ &${\textbf{84.0}}$ \\
\noalign{\hrule height 0.7pt}
 
\end{tabular}
\end{center}
\caption{Ablation study - The  detection performance (AUC (\%)) of different mask strategies on the testing datasets after training on FF++.}
\label{masks}
% \vspace{-0.8cm}
\end{table}

\noindent\textbf{Influence of the masking strategy. }We modify the masking strategies of MAE \cite{imgmae} to improve the generalization. To evaluate the effectiveness of the improved  masking strategy, we compare the proposed masking strategy with masking strategies of MAE and VideoMAE. Furthermore, since the modified strategy randomly selects parts to mask, evaluating the effects of different masked parts is important. To analyze the effectiveness of the ROIs, we compare the proposed strategy with the masking strategy that does not focus on ROIs. We trained models on
the FF++ dataset with different masking strategies.

\indent The results of $1^{st}$, $2^{nd}$, and  $7^{th}$ lines  in  Table \ref{masks} demonstrate that modifying the  masking strategies of MAE \cite{imgmae} and VideoMAE \cite{videomae} can improve the detection performance. The results in the $3^{rd}$, $4^{th}$ and $5^{th}$  lines, which represent methods that mask eye areas, cheek and nose areas, and lip areas, respectively, show a performance degradation compared to the proposed strategy.  That is, random masking a part of all facial parts is more conducive to extracting robust features than masking a certain part only.  Moreover, the results of the $6^{th}$ line and $7^{th}$ lines show that the proposed masking strategy that focuses on ROIs achieves better performance than the  masking strategy without ROIs. The reason  is that the model can better capture the differences between real and fake videos by masking patches in these ROIs, as fake videos typically lack %local and global
consistency. Therefore, the proposed masking strategy is effective in detecting Deepfake videos.

\begin{table}
\setlength{\abovecaptionskip}{0.1cm}
\setlength{\belowcaptionskip}{0.1cm}
\tabcolsep=2pt

\begin{center}

\renewcommand{\tabcolsep}{1.5mm} % enlarge column spacing
\renewcommand{\arraystretch}{1.0}
\begin{tabular}{cccc}
\noalign{\hrule height 0.7pt}
     {}& {CDF}& {DFo}&{DFDC}\\
\noalign{\hrule height 0.7pt}
{\textmd{MAE}}&$76.4$& $88.2$&$71.1$ \\
 {\textmd{VideoMAE}}&$77.4$& $89.3$&$71.8$ \\

 % {\textmd{MAE+Localization stage}}&${{87.2}}$&${{80.0}}$ &${{80.9}}$\\
 % {\textmd{VideoMAE+Localization stage}}&${{88.9}}$&${{73.0}}$ &${{71.9}}$\\

    \textmd{w/o Recovering stage}&$89.0$&$96.5$& $80.9$  \\

\textmd{w/o Localization stage}&$85.8$& $95.4$&$80.0 $\\
 
{\textmd{w/o Meta-learning}}&$89.6$& $96.1$&$81.4$ \\
{\textmd{w/o Mapping}}&$88.2$& $95.3$&$80.9$ \\
{\textmd{w/o Encoder-Decoder}}&$89.9$& $96.7$&$82.8$ \\
  {\textmd{MAE + Localization stage}}&$82.8$& $92.9$&$75.1$ \\
   {\textmd{VideoMAE + Localization stage}}&$83.7$& $93.2$&$75.8$ \\
     {\textmd{RECCE + Localization stage}}&$81.8$& $92.4$&$76.2$ \\ 
 \rowcolor{gray!20}
{\textmd{Delocate}}&${\textbf{91.3}}$&${\textbf{99.1}}$ &${\textbf{84.0}}$  \\
\noalign{\hrule height 0.7pt}
\end{tabular}
\end{center}
\caption{Ablation study - Effects of MAE, VideoMAE, Recovering stage, Localization stage,  Meta-learning, Mapping and Encoder-Decoder.}
\label{twostream}
% \vspace{-0.7cm}
\end{table}

\noindent\textbf{Influence of MAE and VideoMAE.} 
% We adopt the MAE \cite{imgmae} and the vanilla ViT and joint space-time attention of VideoMAE \cite{videomae} for Deepfake detection by modifying the masking strategy and adding the Mapping Network in the second stage. 
We compare the detection performance of the  \textsf{Delocate} with the original MAE and VideoMAE methods for Deepfake detection. The results are shown in the $1^{st}$ and $2^{nd}$ line of Table \ref{twostream}. The detection performance of the original MAE and VideoMAE is lower than that of \textsf{Delocate}, demonstrating the effectiveness of the modifications in \textsf{Delocate}. 

\noindent \textbf{Influence of Recovering stage and Localization stage.}   To validate the performance of each stage, we compare the performance of a single stage with that of both stages combined. The results are shown in the $3^{rd}$ and $4^{th}$ lines of Table \ref{twostream}. We can see that removing either the Recovering stage or the Localization stage degraded the detection performance, as each stage plays a crucial role in Deepfake detection. Combining both stages improves the performance by magnifying the distinction between real and fake videos.
%removing Finetuning Network or Mapping Network can degrade the detection performance. That is because each branch plays a role in Deepfake detection and combining two branches can improve the performance. Specifically, the Finetuning Network can expose inconsistent features by utilizing the encoder of the first stage. The Mapping Network can reveal the inconsistencies by developing the autoencoder of the first stage. The fusion of two branches can magnify the distinction between real videos and fake videos and thus improve the cross-dataset detection performance. 

\noindent \textbf{Influence of Meta-learning, Mapping, and Encoder-Decoder. }We remove the meta-learning, mapping, and Encoder-Decoder module to carry out experiments, respectively, and the results are shown in the $5^{th}$, $6^{th}$, $7^{th}$ line of Table \ref{twostream}. Compared with results of $11^{th}$ line, the method without meta-learning, mapping, and Encoder-Decoder module achieves worse results than the proposed  \textsf{Delocate} with these modules. The meta-learning approach simulates cross-domain detection in the training phase, improving detection performance. The mapping module can reveal the inconsistencies by developing the autoencoder of the Recovering stage, which facilitates the Encoder-Decoder module to locate the forgery regions.  The Encoder-Decoder module achieves the forgery localization, providing a guidance for the classification results. 

\section{Conclusion}
This paper focuses on the detection and localization of Deepfakes, particularly in identifying Deepfake videos with randomly-located tampered traces. By focusing equally on all facial parts rather than relying on specific facial parts, our two-stage model can learn unspecific facial consistencies and general representations.  In the Recovering stage, the model is trained to recover faces from partially masked ROIs on the face, which facilitates the model in learning the facial part consistencies of real videos. In the Localization stage, the model enforces a mapping and an encoder-decoder strategy to expose the forgery areas in synthetic ones.  Extensive experiments illustrate the  generalizability of \textsf{Delocate} in detection and localization.
 \section*{Acknowledgments}
% This work is supported by National Natural Science Foundation
% of China (Grant Nos. U22A2030, 61972142), National Key R\&D
% Program of China (Grant No. 2022YFB3103500),
% This work is supported by  Hunan Provincial Funds for Distinguished Young Scholars (Grant No. 2024JJ2025), Changsha Science and Technology Major
% Project (Grant No. kh2205033), and Changsha Science and Technology Major Project (kh2205033).
This work is supported by National Natural Science Foundation of China (Grant Nos. U22A2030, U20A20174), National Key R\& D Program of China (Grant No. 2022YFB3103500), Hunan Provincial Funds for Distinguished Young Scholars (Grant No. 2024JJ2025). Juan Hu is supported by the Ministry of Education, Singapore, under its MOE AcRF TIER 3 (Grant MOE-MOET32022-0001).  Satoshi Tsutsui is supported by the NTU-PKU Joint Research Institute (a collaboration between the Nanyang Technological University and Peking University that is sponsored by a donation from the Ng Teng Fong Charitable Foundation). Mike Zheng Shou does not receive any funding for this work. 
% \bibliographystyle{named}
% \bibliography{ijcai24}
{\small
\bibliographystyle{named}
\bibliography{ijcai24}
}
\end{document}